\documentclass[letterpaper, 10 pt, conference]{ieeeconf}  % Comment this line out
\IEEEoverridecommandlockouts                              % This command is only
\usepackage[utf8]{inputenc}
\usepackage[T1]{fontenc}

\usepackage{psfrag,epsfig,graphics}
\usepackage{amsmath,amssymb,multirow}
\usepackage{booktabs}
\usepackage{graphicx}
\usepackage{float}
\usepackage{nccmath}%For shorter space after equation 

\usepackage[noadjust]{cite}
\usepackage{multirow}
\usepackage[lined,linesnumbered,ruled]{algorithm2e}

\usepackage{color}  % To include comments in color

\newcommand{\cb}[1]{{\boldsymbol{#1}}}
\newcommand{\cp}[1]{\ifmmode {\mathcal{#1}}\else ${\mathcal{#1}}$\fi}
\usepackage{array}
\newcolumntype{P}[1]{>{\centering\arraybackslash}p{#1}}

\newcommand{\bH}{\boldsymbol{H}}

\newcommand{\bS}{\boldsymbol{S}}

\newcommand{\bn}{\boldsymbol{n}}

\newcommand{\br}{\boldsymbol{r}}
\newcommand{\by}{\boldsymbol{y}}
\newcommand{\bs}{\boldsymbol{s}}
\newcommand{\bu}{\boldsymbol{u}}

\newcommand{\bx}{\boldsymbol{x}}

\newcommand{\calN}{\mathcal{N}}

\newcommand{\bbR}{\mathbb{R}}

\newcommand{\bphi}{\boldsymbol{\phi}}
\newcommand{\bmu}{\boldsymbol{\mu}}

\newcommand{\bSigma}{\boldsymbol{\Sigma}}

\usepackage{color}  % To include comments in color

\definecolor{darkgreen}{rgb}{0., 0.4, 0.}
\definecolor{amber}{rgb}{1.0, 0.49, 0.0}
\definecolor{orange}{rgb}{1.0, 0.4, 0.0}

\title{\LARGE \bf
Cubature Kalman Filter Based Training of Hybrid Differential Equation Recurrent Neural Network Physiological Dynamic Models}

\author{Ahmet Demirkaya$^{1}$, Tales Imbiriba$^{1}$, Kyle Lockwood$^{1}$, Sumientra Rampersad$^{1}$, \\ Elie Alhajjar$^{2}$, Giovanna Guidoboni$^{3}$, Zachary Danziger$^{4}$, Deniz Erdo{\u{g}}mu\c{s}$^{1}$% <-this % stops a space
\thanks{$^{1}$Department of Electrical and Computer Engineering, Northeastern University, Boston, MA 02115, USA; e-mail: \{demirkaya.a, t.imbiriba, lockwood.k, s.rampersad, d.erdogmus\}@northeastern.edu. $^{2}$The United States Military Academy, West Point, NY; e-mail: elie.alhajjar@westpoint.edu. $^{3}$Department of Electrical Engineering and Computer Science, Department of Mathematics, University of Missouri, Columbia, MO; e-mail: guidobonig@missouri.edu. $^{4}$Department of Biomedical Engineering, Florida International University, Miami, FL; e-mail: zdanzige@fiu.edu.}

\thanks {This work was supported in part by NIH (OT2OD030524) and NSF (DMS-1853222/1853303, IIS-1715858, M3X-20040457).
}
}

\begin{document}

\maketitle
\thispagestyle{empty}
\pagestyle{empty}

%%%%%%%%%%%%%%%%%%%%%%%%%%%%%%%%%%%%%%%%%%%%%%%%%%%%%%%%%%%%%%%
\begin{abstract}
Modeling biological dynamical systems is challenging due to the interdependence of different system components, some of which are not fully understood. To fill existing gaps in our ability to mechanistically model physiological systems, we propose to combine neural networks with physics-based models. Specifically, we demonstrate how we can approximate missing ordinary differential equations (ODEs) coupled with known ODEs using Bayesian filtering techniques to train the model parameters and simultaneously estimate dynamic state variables. As a study case we leverage a well-understood model for blood circulation in the human retina and replace one of its core ODEs with a neural network approximation, representing the case where we have incomplete knowledge of the physiological state dynamics. Results demonstrate that state dynamics corresponding to the missing ODEs can be approximated well using a neural network trained using a recursive Bayesian filtering approach in a fashion coupled with the known state dynamic differential equations. This demonstrates that dynamics and impact of missing state variables can be captured through joint state estimation and model parameter estimation within a recursive Bayesian state estimation (RBSE) framework. Results also indicate that this RBSE approach to training the NN parameters yields better outcomes (measurement/state estimation accuracy) than training the neural network with backpropagation through time in the same setting. 
\end{abstract}


%%%%%%%%%%%%%%%%%%%%%%%%%%%%%%%%%%%%%%%%%%%%%%%%%%%%%%%%%%%%%%%
\section{INTRODUCTION}
    
The dynamic behavior of biological systems is often modeled via sets of ordinary differential equations (ODEs) that describe the evolution of biological quantities, such as blood pressure or molecular concentrations, over time. Such models are based on known physics of the described systems and data acquired from experiments with humans, animals, cell cultures, or combinations of such data in multi-scale models. Complex biological systems consist of many interconnecting processes. Generally, the ODEs for some of these processes are not known and it may be difficult to acquire data required to fit model parameters and formulate a mathematical expression underlying the biophysics. Physics-informed machine learning has emerged as a method to fill the gaps in models based on differential equations in general. For example, Raissi et al. developed a deep learning framework to solve problems involving nonlinear partial differential equations~\cite{Raissi2019}, and Weinan et al. used deep learning to solve high-dimensional parabolic equations~\cite{Weinan2017}. While these methods are promising, training a physics-informed deep neural network can be cumbersome due to long cycles and unknown initial states, which reduce the effectiveness of backpropagation through time. Here, we present the use of recursive Bayesian estimation and cubature Kalman filters to efficiently train a physics-informed recurrent neural network. 

Recursive Bayesian estimation is a probabilistic method to estimate a probability density function (PDF) recursively by extracting information about parameters, or \textit{states}, of a dynamical system in real time from measurements of the system output and a system model. The estimate of a state in a given time instant is based on the old estimate of that state and the new measurement data. Kalman filters are recursive Bayesian filters for multivariate normal distributions, which are commonly used in signal processing, navigation and automatic control. The use of Kalman filters to train recurrent neural networks (RNNs) has been proposed and implemented successfully in the past~\cite{Singhal1988,Puskorius1991}. Here we present the first implementation of Kalman-filter-trained RNNs for the estimation of missing ODEs in models of complex biological systems.

We use a validated model for the blood circulation in the human retina to test our method~\cite{guidoboni2014intraocular}. We compare our proposed method to the conventional method, i.e., backpropagation through time (BPTT), for various levels of additive noise. Results show that our method was successful in estimating the system behavior with an NRMSE of 0.038 and was superior to backpropagation for all tested parameters.

%%%%%%%%%%%%%%%%%%%%%%%%%%%%%%%%%%%%%%%%%%%%%%%%%%%%%%%%%%%%%%%

\section{Methods}
\subsection{Cubature Kalman Filter} \label{sec:Cubature}
In this paper we train a hybrid ODE and neural network model using recursive Bayesian state estimation. Specifically, when the predictive and observation models are linear and Gaussian, the state posterior recursion integrals %in~\eqref{eq:chap_kolmo} and~\eqref{eq:evidence} 
can be solved analytically leading to the well-known Kalman time and measurement update equations~\cite{sarkka2013bayesian}. When nonlinear models are in play, required integrals often become intractable and numerical strategies must be sought. Alternatives include linearization of nonlinear functions (extended Kalman filters, EKF) or sampling methods such as particle filters~\cite{arulampalam2002tutorial}, unscented Kalman filters (UKF)~\cite{wan2001unscented}, or cubature Kalman filters (CKF)~\cite{arasaratnam2009cubature}, which assume different levels of system simplicity. 
For instance, EKF, UKF and CKF assume Gaussianity of the measurement and transitional models while handling the integration exploiting this Gaussianity in different ways. While EKF linearizes the models using a first-order Taylor expansion, UKF and CKF use unscented and cubature rules to compute integrals such as

\vspace{-0.03in}

\begin{align}\label{eq:int_l}
    I(\ell) = \int_\cp{D} \ell(\bx) p(\bx) d\bx
\end{align}

where $\ell$ is a nonlinear function of $\bx\in\bbR^{d_x}$ and $p(\bx)=\calN(\bmu, \bSigma)$ is a Gaussian PDF with mean $\bmu$ and covariance $\bSigma$, as a weighted sum of function evaluations of a finite number of deterministic points. For the third-degree cubature rule, the integral in~\eqref{eq:int_l} can be approximated as
\begin{align}\label{eq:cubature_int}
    I(\ell) \approx \frac{1}{2d_x} \sum_{j=1}^{2d_x} \ell(\bS^\top\cb{\xi}_j + \bmu) 
\end{align}
where $\cb{\xi}_j = [\cb{1}]_j \sqrt{2d_x/2}$ are deterministic points~\cite{arasaratnam2009cubature}, and $\bS$ is the lower triangular Cholesky decomposition such that $\bSigma = \bS\bS^\top$. It is important to highlight that the cubature rule demands only two points per dimension of $\bx$, i.e., $2d_x$, to evaluate the sum in~\eqref{eq:cubature_int}, making it more suitable when working in high-dimensional state-spaces. Assuming Gaussianity of state posteriors, CKF can solve the integrals required in the Bayesian recursion as well as the moments (mean and covariance) of the new state posterior~\cite{arasaratnam2009cubature}.

In contrast, particle filters do not assume any particular distribution, instead approximating the distribution as a linear combination of Dirac deltas. Thus, moments of propagated particles can be easily computed. One drawback of particle filters is the high number of particles needed to accurately represent distributions. This issue is profoundly aggravated if the state-space dimension is large, making this filtering strategy unfeasible in such scenarios~\cite{imbiriba_gppf_2020}.
    
\subsection{Training Recurrent Neural Networks with Augmented Latent Space}
Bayesian filtering (BF) has been considered to train neural networks applied to dynamical systems~\cite{haykin2004kalman}, 
 using EKF~\cite{erdogmus2002modified} and more recently applied to target tracking in indoor navigation \cite{wu2019wifi}, where the state transition and measurement models can be arbitrary feed-forward neural networks such as:

 \vspace{-0.07in}

\begin{align}
    \dot{\bs} &= g(\bs, \bu; \tilde{\cb{\omega}}) + \tilde{\bn} \nonumber\\
    \by &= q(\bs; \cb{\phi}) + \br
\end{align}

where $\tilde{\bn} \sim \calN (0,\sigma_{\tilde{n}}^2\cb{I})$, $\bs$ is the vector of latent states, $\bu$ are the external inputs, $\br$ is the additive measurement noise with $\br \sim \calN (0,\sigma_{\br}^2\cb{I})$, and $\tilde{\cb{\omega}}$ and $\cb{\phi}$ are the model parameters of $g$ and $q$, respectively. 

The discretized model can be obtained by approximating the derivative via finite differences leading to
% \begin{align}
%     % \bs_{k} &= \bs_{k-1} + \Delta_t g(\bs_{k-1}; \tilde{\cb{\omega}}_{k-1}) + \Delta_t\tilde{\bn}_{k-1} \nonumber\\
%     \bs_{k} &= \bs_{k-1} + \Delta_t g(\bs_{k-1}, \bu_{k-1}; \tilde{\cb{\omega}}) + \Delta_t\tilde{\bn}_{k-1} \nonumber\\
%     \by_k &= q(\bs_k; \cb{\phi}) + \br_k
% \end{align}
% which can be rewritten as 
\vspace{-0.07in}

\begin{align}
    % \bs_{k} &= \bs_{k-1} + g(\bs_{k-1};\cb{\omega}_{k-1}) + \bn_{k-1} \nonumber \\
    \bs_{k} &= \bs_{k-1} + g(\bs_{k-1}, \bu_{k-1};\cb{\omega}) + \bn_{k-1} \nonumber \\
    \by_k &= q(\bs_k; \cb{\phi}) + \br_k. \label{eq: discretizedModel}
\end{align}
where we assumed that the time-step $\Delta_t$ is incorporated in the NN weights, $\bn_k$ is process noise with $\bn_k \sim \calN (0, \sigma_n^2\cb{I})$, where $\sigma_n = {\Delta_t}\sigma_{\tilde{n}}$, and $\br_k$ is the additive measurement noise with $\br_k \sim \calN (0,\sigma_{\br}^2\cb{I})$.

In order to learn the model parameters using a BF approach, we augment the latent space as:
% The augmented latent space becomes:
\vspace{-0.09in}
\begin{align}\label{eq:discretizedModel2}
    \cb{\omega}_{k} &= \cb{\omega}_{k-1} \nonumber\\
    \bphi_{k} &= \bphi_{k-1} \nonumber\\
    \bs_{k} &= \bs_{k-1} + g(\bs_{k-1}, \bu_{k-1}; \cb{\omega}_{k-1}) + \bn_{k-1} \\
    \by_k &= q(\bs_k; \cb{\phi}_k) + \br_k \nonumber
\end{align}
which could be rewritten as 
\vspace{-0.03in}
\begin{align}\label{eq:compact_model}
    \bx_{k} &= f(\bx_{k-1}, \bu_{k-1}) + \cb{\varepsilon}_{k-1} \nonumber\\
    \by_k &= h(\bx_k) + \br_k
\end{align}
where $\bx_k = [\cb{\omega}_{k}^\top, \bphi_{k}^\top, \bs_k^\top]^\top$, $f (\bx_{k-1}, \bu_{k-1})=[\cb{\omega}_{k-1},\bphi_{k-1}, g(\bs_{k-1}, \bu_{k-1}; \cb{\omega}_{k-1})]^\top$, and $h(\bx_k) = q(\bs_k; \bphi_{k})$.
The above model can then be tackled using filtering techniques~\cite{erdogmus2002modified, wu2019wifi} as discussed in Section~\ref{sec:Cubature}.

% %%%%%%%%%%%%%%%%%%%%%%%%%%%%%%%%%%%%%%%%%%%%%%%%%%%%%%%%%%%%%%%

\subsection{Application to the Retinal Circulation}\label{sec:retina_model}
The proposed approach was tested on a validated model for the retinal circulation proposed by Guidoboni et al. ~\cite{guidoboni2014intraocular}. The model consists of four nonlinear ODEs describing 
the blood flow through the retinal vasculature and the central retinal vessels. Leveraging the electric analogy to fluid flow, electric currents and electric potentials represent blood flow rates and blood pressures ($P$), while resistors and capacitors represent hydraulic resistance and vessel compliance~\cite{sacco2019comprehensive}. The model includes five vascular compartments, whose internal pressures are denoted from 1 to 5: central retinal artery (CRA): $P_1$; arterioles: $P_2$; capillaries: $P_3$; venules: $P_4$; central retinal vein (CRV): $P_5$. Pressures $P_{in}$ and $P_{out}$ are inlet and outlet pressures that drive the network. By applying Kirchhoff's laws of currents and voltages and leveraging the constitutive equations of each resistor and capacitor, the following ODE system is obtained:  
\begin{equation}
\dot{P_1} = \mathcal F_1(P_1,P_2;P_{in})\\
\label{eq:p1}
\end{equation}

\useshortskip
\begin{equation}
\dot{P_2} = \mathcal F_2(P_1,P_2,P_4)
\label{eq:p2}
\end{equation}

\useshortskip
\begin{equation}
\dot{P_4} = \mathcal F_4(P_2,P_4,P_5)
\label{eq:p4}
\end{equation}

\useshortskip
\begin{equation}
\dot{P_5} = \mathcal F_5(P_4,P_5;P_{out})
\label{eq:p5}
\end{equation}
The explicit expressions of the functional $\mathcal F_i $ representing the time rate of change of the pressures $P_i$ can be found in~\cite{guidoboni2014intraocular}.
%is used to calculate the derivative of the pressure measured at point $P_i $.
To test our proposed method, we treat $P_4$ as though its dynamics were unknown and replace $\mathcal F_4$ in equation (15) with a neural network with a built-in integrator as follows
\begin{equation}
\dot{P_4} = q_4(P_2,P_4,P_5; \cb{\omega}),
\label{eq:nn}
\end{equation}
where $q_4$ is a feed-forward neural network with parameters $\cb{\omega}$, while equations~\eqref{eq:p1},~\eqref{eq:p2} and~\eqref{eq:p5} are used for the dynamics of pressures $P_1$, $P_2$ and $P_5$. Note that equations \eqref{eq:p1}--\eqref{eq:p5} have recursive relations that become apparent when time is discretized: $\bs_{k+1} = \bs_k + \Delta_t \mathbf{f}(\bs_k,\bu_k)$, where $\bs$ consists of four pressure values $P_i$, and $\mathbf{f}$ represents the functions on the right side of equations~\eqref{eq:p1},~\eqref{eq:p2},~\eqref{eq:nn} and \eqref{eq:p5}. Therefore, the neural network given in equation \eqref{eq:nn} is trained using an RNN that encapsulates this discrete-time approximation for integration over time. In the RNN, state variables are used to represent the pressures in each retinal vascular compartment. %The neural network in equation \eqref{eq:nn} is used for estimation of the latent state and the predefined equations in \eqref{eq:p1}, \eqref{eq:p2} and \eqref{eq:p5} are used for explicitly defined states.
Both backpropagation through time and cubature Kalman filters can be used as training methodologies. In the next section, their performances are compared. %Throughout the paper, a neural network with a fixed number of parameters is used for both BPTT and CKF training.
The feed-forward neural network $q_4$ is encapsulated in the RNN that couples known differential equations with the NN model for the unknown dynamics. The $q_4$ used in the following numerical analyses of both BPTT and CKF training consists of a single hidden layer with 20 hidden units and a total of 103 parameters. For the BPTT training, we used Adam optimizer with a learning rate of $5 \times 10^{-3}$. 

For the models in~\eqref{eq:compact_model}, we set the vector of latent states as discretized pressure values at time $k$, i.e., $\bs_k = [P_{1,k},\ldots, P_{5,k}]^\top$, $g(\bs_k;\cb{\omega}_k)= [\mathcal{F}_{1}(P_{1,k},P_{2,k};P_{in,k}), \; \mathcal {F}_{2}(P_{1,k},P_{2,k},P_{4,k}), \; \, q_4(P_{2,k}, P_{4,k}, \newline P_{5,k};\cb{\omega}_k),\allowbreak \mathcal{F}_{5}(P_{4,k}, P_{5,k};P_{out,k})]$, and $h(\bs_k) = \bH\bs_k$, where $\bH$ is a matrix with zeros and ones selecting measured pressures and blocking unobservable states. For credible practices of modeling and simulation, we followed "Ten Simple Rules" suggested by Erdemir et. al \cite{erdemir_credible_2020}.

\vspace{-0.25cm}

%%%%%%%%%%%%%%%%%%%%%%%%%%%%%%%%%%%%%%%%%%%%%%%%%%%%%%%%%%%%%%%
\section{RESULTS}\label{sec:results}
We used the retinal circulation model from \cite{guidoboni2014intraocular} to demonstrate the effectiveness of CKF training of a hybrid ODE and RNN model. In order to compare the performance of CKF and BPTT training, we performed Monte Carlo simulations with 100 runs each, with random initialization of the model parameters at each run. As error metrics we consider the mean absolute percentage error (MAPE) and normalized root-mean-square error (NRMSE), which are computed as $\text{MAPE} = 100 d_y^{-1} \allowbreak\sum_{i=1}^{d_y}\left(\sum_{k=1}^{n_t}|\hat{y}_{i,k}-y_{i,k}|/\sum_{k=1}^{n_t} |y_{i,k}| \right)$ and  $\text{NRMSE} = \allowbreak \{\sum_{i=1}^{d_y} [\sum_{k=1}^{n_t}(\hat{y}_{i,k}-y_{i,k})^2/(n_t (y^{\max}_i-y^{\min}_i))]/d_y\}^{1/2}$, respectively, with $\by_k = [P_{1,k}, \ldots, P_{5,k}]^\top \in \bbR^{d_y}$, and $n_t$ is the total number of time samples used.

Pressures $P_{in}$ and $P_{out}$ were used as inputs, $P_1$, $P_2$ and $P_5$ were used as observable states to train the model and $ P_4$ was assumed to be not observable and modeled as a latent space. To take potential noise in the system into account, we introduced white noise to all inputs and observable states in our experiments. We used five different noise levels with signal-to-noise ratios (SNR) given in Table \ref{table:errors}, where resulting quantitative error metrics are also reported. Both MAPE and NRMSE increased with decreasing SNR for CKF as well as BPTT. Across different noise levels, the errors for BPTT were 1.9--5.4 times higher than for CKF. 

Figs. 1--3 show the \textit{ground truth} values, i.e., values obtained using the ODEs provided by Guidoboni et al. ~\cite{guidoboni2014intraocular}, and estimations of the observable and unobserved states after training the model with noisy pressure values. Fig. 1 presents a single simulation for the CKF-trained model, demonstrating the noise on the inputs and estimated values. Despite the existence of noise, the estimated values converge quickly to the ground truth values. Figs. 2 and 3 demonstrate the mean and 95\% confidence intervals for the Monte Carlo simulations using CKF (Fig. 2) and BPTT (Fig. 3) training for simulations with an SNR of 22.56. For both methods, the estimated values converge to the ground truth values. However, the CKF-trained model is able to estimate the unobserved state in fewer steps with higher confidence than the same model trained with BPTT. Fig.~\ref{fig:noise_levels} presents learning curves for both methods, showing that CKF training reaches an error below the noise level faster and converge in fewer steps than BPTT.
\vspace{-0.4cm}

\begin{figure}[htb]
\centerline{\includegraphics[trim={0 0 0 0.0cm},clip,width=\columnwidth]{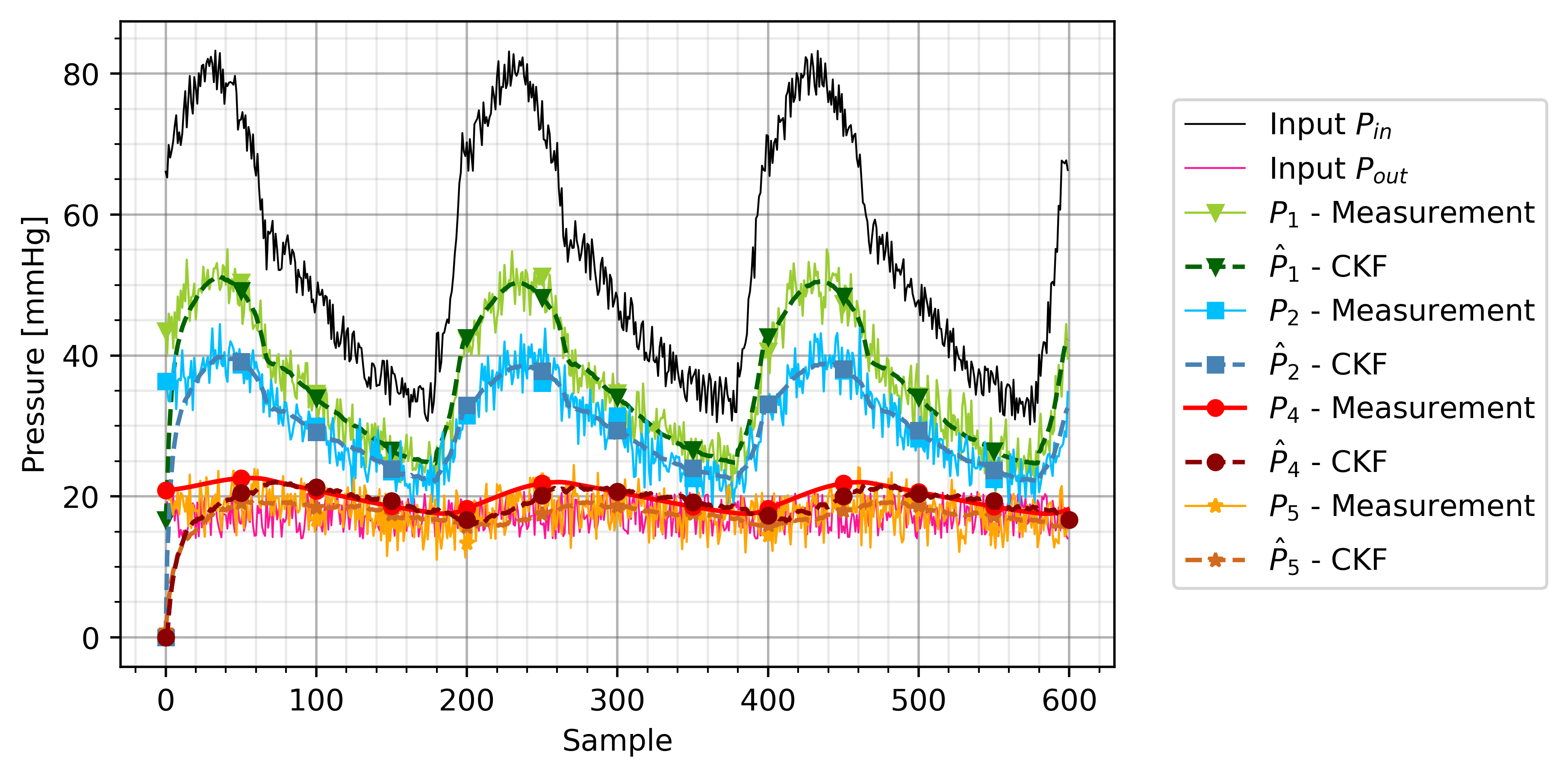}}
\vspace{-0.3cm}
\caption{An example simulation for the CKF-trained model with white noise (SNR of 22.56) added to the inputs and observable states.}
\label{fig:single_result}
\end{figure}

\vspace{-0.35cm}

\begin{figure}[htb]
\vspace{-0.35cm}
\centerline{\includegraphics[trim={0 0 0 0.1cm},clip,width=\columnwidth]{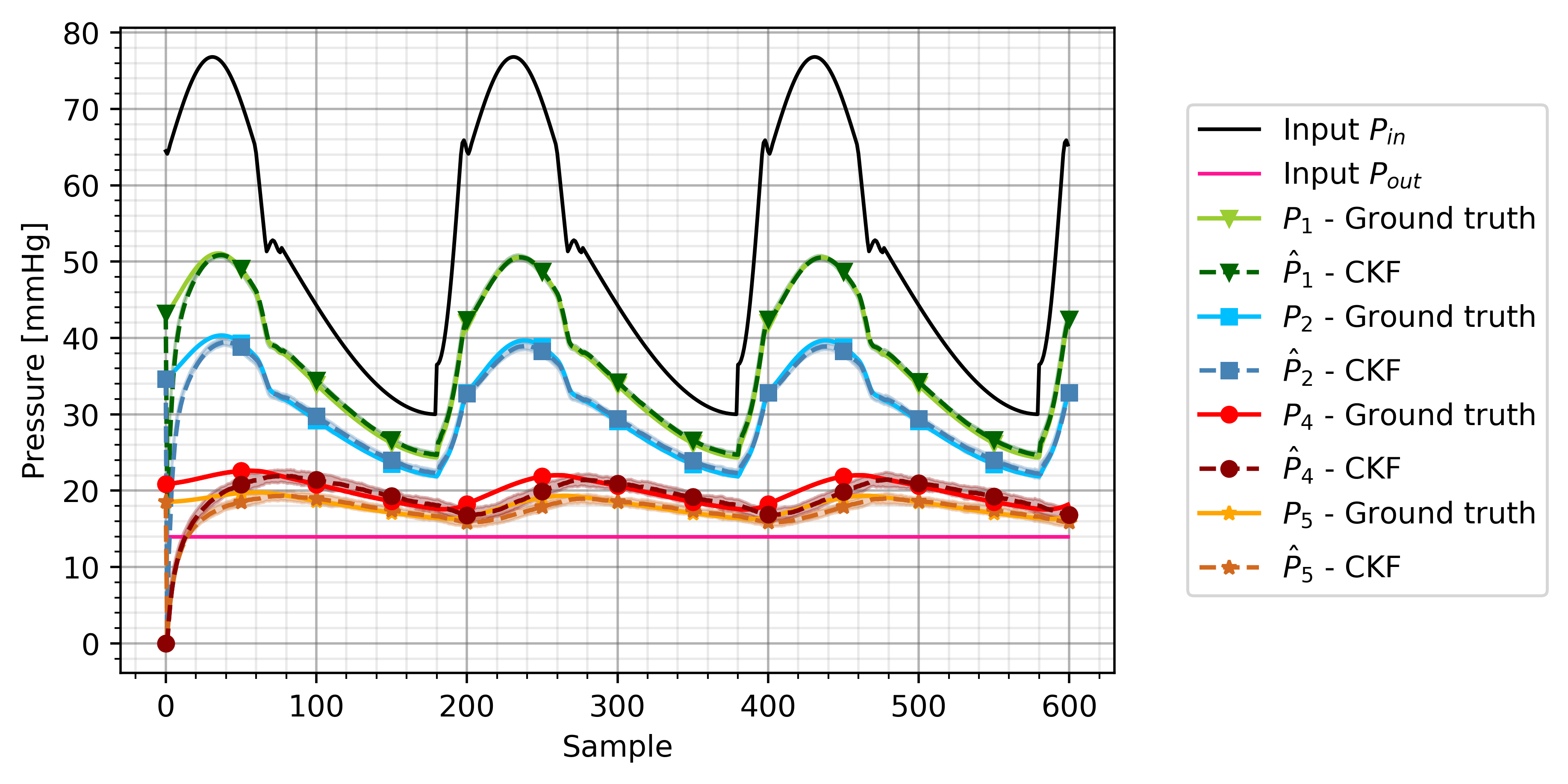}}  
\vspace{-0.4cm}
\caption{Result of CKF training on 100 Monte Carlo simulations with white noise (SNR of 22.56). Ground truth values (solid lines) and estimated values (dashed lines) are shown, with 95\% confidence intervals for all estimated states.}
\label{fig:monte_carlo_ckf}
\vspace{-0.15cm}
\end{figure}

\begin{figure}[htb]
\vspace{-0.65cm}
\centerline{\includegraphics[trim={0 0 0 0},clip,width=\columnwidth]{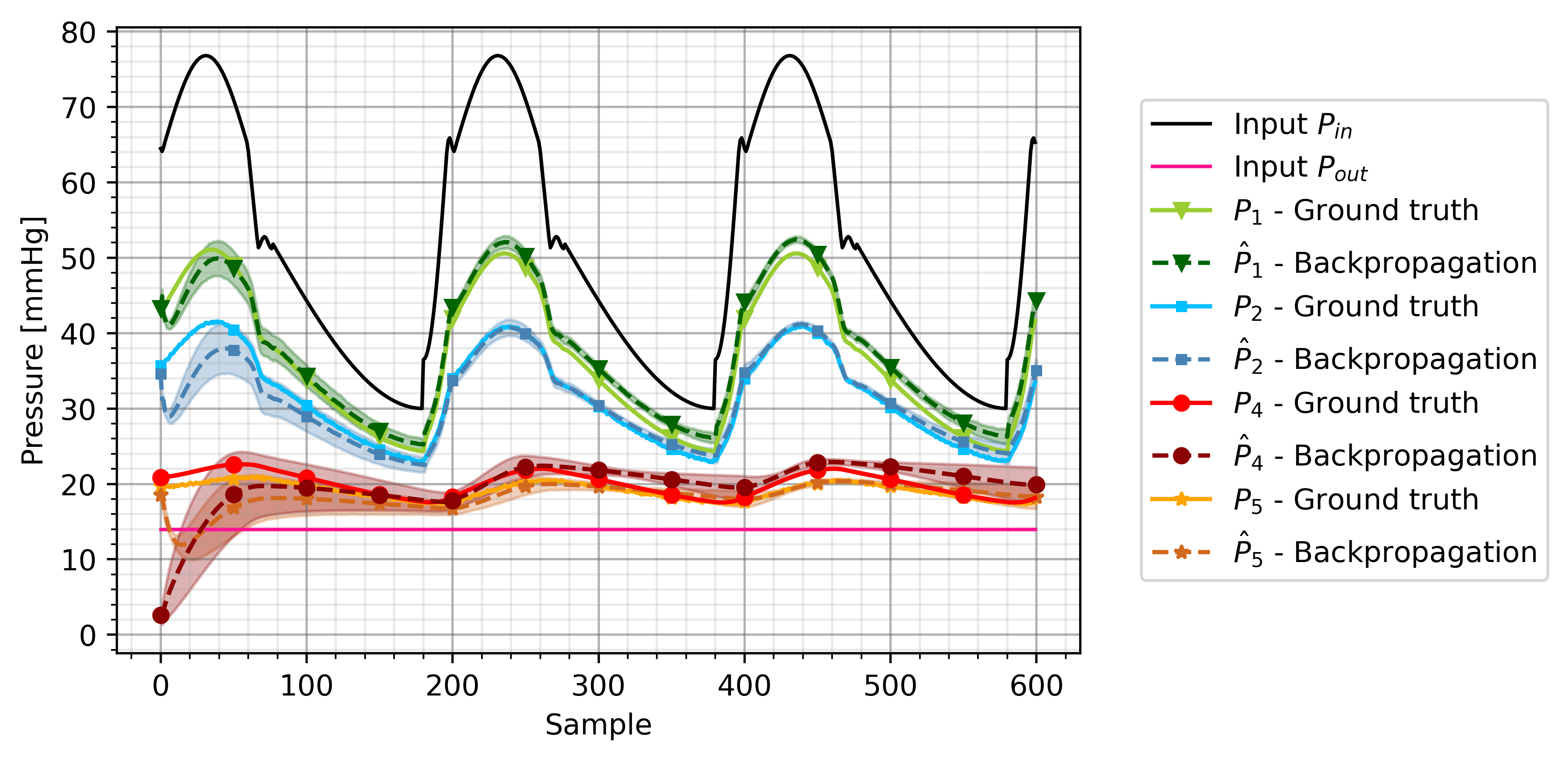}}  
\vspace{-0.3cm}
\caption{Result of BPTT training on 100 Monte Carlo simulations with white noise (SNR of 22.56). Ground truth values (solid lines) and estimated values (dashed lines) are shown, with 95\% confidence intervals for all estimated states.}
\label{fig:mote_carlo_backprop}
\end{figure}

\vspace{-0.15cm}

\begin{figure}[htb]
\centerline{\includegraphics[trim={0 0 0 0.0cm},clip,width=0.8\columnwidth]{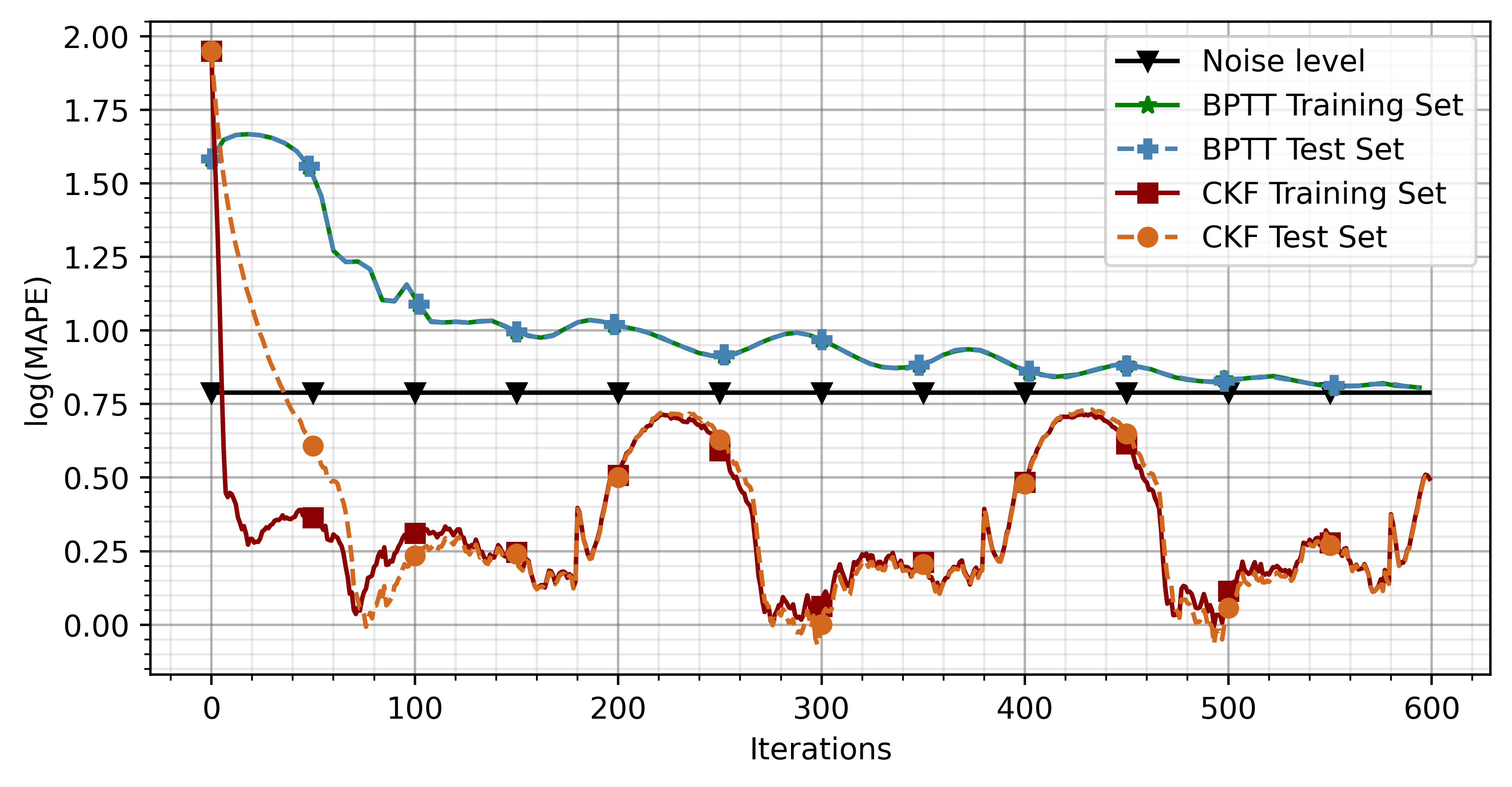}}
\vspace{-0.13in}
\caption{Learning curves of CKF and BPTT training for a simulation with white noise (SNR of 22.56). Results on the training set (solid lines) and test set (dashed lines) are shown in comparison with the noise level (black solid line).}
\label{fig:noise_levels}
\vspace{-0.15cm}
\end{figure}

\vspace{-0.2cm}

\begin{table}[htb]
\vspace{-0.15in}

\caption{MAPE and NRMSE averaged across 100 Monte Carlo experiments}
\vspace{-0.07in}

\begin{tabular}{|P{0.8cm}|P{1.4cm}|P{1.4cm}|P{1.4cm}|P{1.4cm}|}
\hline
\textbf{SNR} & \textbf{CKF MAPE} & \textbf{BPTT MAPE}& \textbf{CKF NRMSE} & \textbf{BPTT NRMSE} \\ \hline
49.53  & \textbf{1.47}  & 5.36 & \textbf{0.038} & 0.204 \\ \hline
39.52  & \textbf{1.82}  & 5.85 & \textbf{0.044} & 0.206 \\ \hline
32.58  & \textbf{2.32}  & 5.74 & \textbf{0.054} & 0.207 \\ \hline
29.51  & \textbf{2.59}  & 5.92 & \textbf{0.061} & 0.208 \\ \hline
22.56  & \textbf{3.54}  & 6.61 & \textbf{0.093} & 0.208 \\ \hline
\end{tabular}
\label{table:errors}
\vspace{-0.3in}

\end{table}

\vspace{-0.4cm}
%%%%%%%%%%%%%%%%%%%%%%%%%%%%%%%%%%%%%%%%%%%%%%%%%%%%%%%%%%%%%%%
\section{Discussion}\label{sec:discussion}

The experimental results indicate that RNN models can fill modeling gaps in ODE-based physiological models given that the set of inputs and outputs of the model can be observed. In the retinal circulation model we obtained error metrics in terms of mean absolute value reaching thresholds below the noise floor. 
When comparing CKF and BPTT results, we observed that CKF provides significant improvement for both training convergence and test performance as can be attested by the results presented in Fig.~\ref{fig:noise_levels} and Table~\ref{table:errors}.
This is due to the fact that BF strategies keep updating the states continuously whenever a new measurement is available. Although we opted to stop RNN parameter updates during the testing phase, in practice we could keep updating the parameters indefinitely, allowing the model to possibly adapt to physiological states not contained in the training set.   

A common problem in solving ODEs is how to determine the initial conditions. This issue persists in the hybrid ODE-RNN scheme discussed in this paper, especially for the model trained with BPTT. In theory, this could be solved by including the initial states as extra model parameters that need to be estimated in both training and testing phases. However, when testing this approach we were never able to obtain accurate estimations for the initial physiological states. On the other hand, by adopting the BF approach we changed the problem's perspective from \emph{estimating the previous state} with BPTT to \emph{estimating the current state given the past set of measurements} with CKF. Thus, in the BF formulation we may disregard the initial states since the filtering updates quickly move the states near the correct value. This difference is key in modeling complex systems since hundreds or even thousands of initial state conditions might have to be found. 

The implementation aspects of the BF strategy is also important, since nonlinear systems often lead to numerical approaches that scale poorly with the dimension of the state space. Here, we considered third-degree cubature rules to solve the filtering integrals under the assumption of Gaussianity of the states. Cubature rules are particularly interesting since the number of required particles are relatively small (twice the state-space dimension) and deterministic, and the solution is exact for monomials of degree three or less~\cite{arasaratnam2009cubature}. Propagating such particles through the state transition and observation models, however, leads to multiple independent model evaluations, which drastically increases the computational cost of the optimization procedure during the training phase. Nevertheless, independent evaluation of such particles allows for easy parallelization of the algorithm.

\vspace{-0.07in}

%%%%%%%%%%%%%%%%%%%%%%%%%%%%%%%%%%%%%%%%%%%%%%%%%%%%%%%%%%%%%%%
\section{CONCLUSIONS}\label{sec:conclusion}

In this paper we presented a strategy to complement ODE-based models with RNNs, where we compared two different learning methodologies: backpropagation and Bayesian filtering with CKF. Using a validated human retinal circulation model, we showed that both methodologies were capable of approximating the missing ODE, and that CKF greatly outperformed BPTT due to the time updates of latent states even during the testing phase. These results show that BF-based strategies hold great potential for approximating the dynamics of unknown states in ODE systems. 

%\vspace{-0.8in}
\vspace{0.1cm}

%%%%%%%%%%%%%%%%%%%%%%%%%%%%%%%%%%%%%%%%%%%%%%%%%%%%%%%%%%%%%%%
% \section*{ACKNOWLEDGMENT}
%     The preferred spelling of the word ``acknowledgment'' in America is without an ``e'' after the ``g''. Avoid the stilted expression, ``One of us (R. B. G.) thanks . . .''  Instead, try ``R. B. G. thanks''. Put sponsor acknowledgments in the unnumbered footnote on the first page.
%%%%%%%%%%%%%%%%%%%%%%%%%%%%%%%%%%%%%%%%%%%%%%%%%%%%%%%%%%%%%%%
\bibliographystyle{IEEEtran}
\bibliography{root}

\begin{thebibliography}{10}
\providecommand{\url}[1]{#1}
\csname url@samestyle\endcsname
\providecommand{\newblock}{\relax}
\providecommand{\bibinfo}[2]{#2}
\providecommand{\BIBentrySTDinterwordspacing}{\spaceskip=0pt\relax}
\providecommand{\BIBentryALTinterwordstretchfactor}{4}
\providecommand{\BIBentryALTinterwordspacing}{\spaceskip=\fontdimen2\font plus
\BIBentryALTinterwordstretchfactor\fontdimen3\font minus
  \fontdimen4\font\relax}
\providecommand{\BIBforeignlanguage}[2]{{%
\expandafter\ifx\csname l@#1\endcsname\relax
\typeout{** WARNING: IEEEtran.bst: No hyphenation pattern has been}%
\typeout{** loaded for the language `#1'. Using the pattern for}%
\typeout{** the default language instead.}%
\else
\language=\csname l@#1\endcsname
\fi
#2}}
\providecommand{\BIBdecl}{\relax}
\BIBdecl

\bibitem{Raissi2019}
\BIBentryALTinterwordspacing
M.~Raissi, P.~Perdikaris, and G.~Karniadakis, ``Physics-informed neural
  networks: A deep learning framework for solving forward and inverse problems
  involving nonlinear partial differential equations,'' \emph{Journal of
  Computational Physics}, vol. 378, pp. 686--707, 2019. [Online]. Available:
  \url{https://www.sciencedirect.com/science/article/pii/S0021999118307125}
\BIBentrySTDinterwordspacing

\bibitem{Weinan2017}
\BIBentryALTinterwordspacing
W.~E, J.~Han, and A.~Jentzen, ``Deep learning-based numerical methods for
  high-dimensional parabolic partial differential equations and backward
  stochastic differential equations,'' \emph{Communications in Mathematics and
  Statistics}, vol.~5, no.~4, p. 349–380, Nov 2017. [Online]. Available:
  \url{http://dx.doi.org/10.1007/s40304-017-0117-6}
\BIBentrySTDinterwordspacing

\bibitem{Singhal1988}
S.~Singhal and L.~Wu, ``Training multilayer perceptrons with the extended
  {K}alman algorithm,'' in \emph{Advances in Neural Information Processing
  Systems}, 1988, pp. 133--140.

\bibitem{Puskorius1991}
G.~Puskorius and L.~Feldkamp, ``Decoupled extended {K}alman filter training of
  feedforward layered networks,'' in \emph{Proceedings of the International
  Joint Conference on Neural Networks}, 1991, pp. 771--777.

\bibitem{guidoboni2014intraocular}
G.~Guidoboni, A.~Harris, S.~Cassani, J.~Arciero, B.~Siesky, A.~Amireskandari,
  L.~Tobe, P.~Egan, I.~Januleviciene, and J.~Park, ``Intraocular pressure,
  blood pressure, and retinal blood flow autoregulation: a mathematical model
  to clarify their relationship and clinical relevance,'' \emph{Investigative
  Ophthalmology \& Visual Science}, vol.~55, no.~7, pp. 4105--4118, 2014.

\bibitem{sarkka2013bayesian}
S.~S{\"a}rkk{\"a}, \emph{Bayesian filtering and smoothing}.\hskip 1em plus
  0.5em minus 0.4em\relax Cambridge University Press, 2013, no.~3.

\bibitem{arulampalam2002tutorial}
M.~S. Arulampalam, S.~Maskell, N.~Gordon, and T.~Clapp, ``A tutorial on
  particle filters for online nonlinear/non-gaussian bayesian tracking,''
  \emph{IEEE Transactions on signal processing}, vol.~50, no.~2, pp. 174--188,
  2002.

\bibitem{wan2001unscented}
E.~A. Wan, R.~Van Der~Merwe, and S.~Haykin, ``The unscented {K}alman filter,''
  \emph{Kalman filtering and neural networks}, vol.~5, no. 2007, pp. 221--280,
  2001.

\bibitem{arasaratnam2009cubature}
I.~Arasaratnam and S.~Haykin, ``Cubature {K}alman filters,'' \emph{IEEE
  Transactions on automatic control}, vol.~54, no.~6, pp. 1254--1269, 2009.

\bibitem{imbiriba_gppf_2020}
T.~Imbiriba and P.~Closas, ``Enhancing particle filtering using gaussian
  processes,'' in \emph{2020 IEEE 23rd International Conference on Information
  Fusion (FUSION)}.\hskip 1em plus 0.5em minus 0.4em\relax IEEE, 2020, pp.
  1--7.

\bibitem{haykin2004kalman}
S.~Haykin, \emph{Kalman filtering and neural networks}.\hskip 1em plus 0.5em
  minus 0.4em\relax John Wiley \& Sons, 2004, vol.~47.

\bibitem{erdogmus2002modified}
D.~Erdogmus, J.~C. Sanchez, and J.~C. Principe, ``Modified {K}alman filter
  based method for training state-recurrent multilayer perceptrons,'' in
  \emph{Proceedings of the 12th IEEE Workshop on Neural Networks for Signal
  Processing}.\hskip 1em plus 0.5em minus 0.4em\relax IEEE, 2002, pp. 219--228.

\bibitem{wu2019wifi}
P.~Wu, T.~Imbiriba, G.~LaMountain, J.~Vil{\`a}-Valls, and P.~Closas, ``Wifi
  fingerprinting and tracking using neural networks,'' in \emph{Proceedings of
  the 32nd International Technical Meeting of the Satellite Division of The
  Institute of Navigation (ION GNSS+ 2019)}, 2019, pp. 2314--2324.

\bibitem{sacco2019comprehensive}
R.~Sacco, G.~Guidoboni, and A.~Mauri, \emph{A comprehensive physically-based
  approach to modeling in bioengineering and life sciences}.\hskip 1em plus
  0.5em minus 0.4em\relax Academic Press. Elsevier, 2019, iSBN: 9780128125182.

\bibitem{erdemir_credible_2020}
A.~Erdemir, L.~Mulugeta, J.~P. Ku, A.~Drach, M.~Horner, T.~M. Morrison,
  G.~C.~Y. Peng, R.~Vadigepalli, W.~W. Lytton, and J.~G. Myers,
  ``\BIBforeignlanguage{en}{Credible practice of modeling and simulation in
  healthcare: ten rules from a multidisciplinary perspective},''
  \emph{\BIBforeignlanguage{en}{Journal of Translational Medicine}}, vol.~18,
  no.~1, p. 369, Dec. 2020.

\end{thebibliography}

\end{document}